\newcommand\SEKIusersusepackages
{
\usepackage{mathpartir}
\usepackage{amsmath}
\usepackage{amssymb}
\usepackage{xspace}
\usepackage{url}
\usepackage{paralist}
\usepackage{subfigure}
\usepackage{wrapfig}
\usepackage{color}
\definecolor{lightgray}{rgb}{0.75,0.75,0.75}
}

\newcommand\SEKIREVIEWSTATUS{1}

\newcommand\SEKISUBEDITOR
{Erica Melis 
\\FR Informatik, Universit\"at des Saarlandes, D--66123 Saarbr\"ucken, Germany
\\E-mail: {\tt melis@activemath.org}
\\WWW: \url{http://www.activemath.org/~melis}
}

\newcommand\SEKIREVIEWER
{Claus-Peter Wirth 
\\E-mail: {\tt wirth@logic.at}
\\WWW: \url{http://www.ags.uni-sb.de/~cp}
}

\newcommand\SEKIREFEREEPROCESS
{{\em Granularity-Adaptive Proof Presentation.}
  Proceedings of the 14th International Conference on Artificial Intelligence in Education;
  Brighton, UK, 
  2009. 
  Submitted.
}

\newcommand\SEKIDOCUMENTCLASS{0}

\newcommand\SEKIYEAR{2009}

\newcommand\SEKINUMBEROFYEAR{01}

\newcommand\SEKITYPE{1}

\newcommand\SEKIPUBLISHEDTYPE{3}
\newcommand\SEKIPUBLISHEDAS
{{\em Granularity-Adaptive Proof Presentation.}
  Proceedings of the 14th International Conference on Artificial Intelligence in Education;
  Brighton, UK, 
  2009. 
  Submitted.}
\title
{Granularity-Adaptive Proof Presentation
}

\author
{Marvin Schiller
\\{\footnotesize
German Research Center for Artificial Intelligence (DFKI), Bremen, Germany}
\\{\footnotesize\url{Marvin.Schiller@dfki.de}}
\\[2ex]
Christoph Benzm\"uller
\\{\footnotesize International University in Germany, Bruchsal, Germany}
\\{\footnotesize\url{c.benzmueller@googlemail.com}}}

\date{\small
\SEKIedition\\
Submitted February\,2, 2009\\
February\,27, 2009
}

\input seki-deckblatt-3

\newcommand\figurebert[1]{\begin{figure*}[#1] \small
\begin{minipage}{1\textwidth} 
\steplabelA{1} Let $x$ be an element of $A\cap (B \cup C)$, \steplabelA{2} then $x \in A$ and $x \in B
\cup C$. \steplabelA{3} This means that $x \in A$, and either $x \in B$ or $x \in C$. \steplabelA{4}
Hence we either have (i) $x \in A$ and $x \in B$, or we have (ii) $x \in A$
and $x \in C$. \steplabelA{5} Therefore, either $x \in A\cap B$ or $x \in A\cap C$, so 
\steplabelA{6} $x \in (A\cap B)\cup (A\cap C)$. \steplabelA{7} This shows that $A \cap (B \cup  C)$ is a
subset of $(A \cap B) \cup  (A \cap C)$. \steplabelA{8} Conversely, let y be an element
of $(A\cap B)\cup (A\cap C)$. \steplabelA{9} Then, either (iii) $y \in A\cap B$, or (iv)
$y \in A \cap C$. \steplabelA{10} It follows that $y \in A$, and either $y \in B$ or $y
\in C$. \steplabelA{11} Therefore, $y \in A$ and $y \in B \cup  C$ so that $y \in A \cap
(B \cup  C)$. \steplabelA{12} Hence $(A\cap B) \cup  (A\cap C)$ is a subset of $A \cap (B
\cup  C)$. \steplabelA{13} In view of Definition 1.1.1, we conclude that the sets $A \cap (B \cup  C)$ and $(A\cap B) \cup  (A\cap C)$ are equal.
\end{minipage}
\caption{Proof of the statement $A \cap (B \cup  C)$ = $(A\cap B) \cup
  (A\cap C)$, reproduced from \cite{BartleSherbert} 
}
\label{bartlesherbertproof}
\vspace{1.5em}
\end{figure*}}

\newcommand\figurealproof[1]{\begin{figure*}[#1]\scalebox{1.3}{
\tiny
\begin{minipage}{0.75\textwidth}
\begin{center}
\begin{mathpar}
\mprset{sep=.1em} 
\inferrule*[Left={\scriptsize Def eq (1)}]{
  \inferrule*[Left={\scriptsize Def$\subseteq$ (2)}]{\inferrule*[Left={\scriptsize Def$\cap$
      (3)}]{\inferrule*[Left={\scriptsize Def$\cup$
        (4)}]{\inferrule*[Left={\scriptsize distr(5)}]{\inferrule*[Left={\scriptsize Def$\cap$
            (6)}]{\inferrule*[Left={\scriptsize Def$\cap$ (7)}]{\inferrule*[Left={\scriptsize Def$\cup$ (8)}]{\inferrule*{}{x\in\!\textbf{S} \vdash x\in\!\textbf{S}} }{(x\!\in\!(A\!\cap\!B)\!\vee\!x\!\in\!(A\!\cap\!C))
              \vdash x\!\in\!\textbf{S}} }{(x\!\in\!(A\!\cap\!B)
            \vee x\!\in\!A\!\wedge\!x\!\in\!C) \vdash x\!\in\!\textbf{S}} }{(x\!\in\!A\!\wedge\!x\!\in\!B \vee x\!\in\!A\!\wedge x\!\in C) \vdash
          x\!\in\!\textbf{S}} }{(x\!\in\!A\!\wedge\!(x\!\in\!B \vee
        x\!\in\!C)) \vdash x\!\in\!\textbf{S}} }{(x\!\in A\!\wedge
      x\!\in\!(B\!\cup\!C)) \vdash x\!\in\!\textbf{S}} }{(x\!\in\!(A
    \cap (B\!\cup\!C))) \vdash x\!\in\!\textbf{S}} }{\vdash (A\!
  \cap (B\!\cup\!C))\!\subseteq\!\textbf{S}} \\ \inferrule*[Right={\scriptsize Def$\subseteq$ (9)}]{\inferrule*[Right={\scriptsize Def$\cup$
    (10)}]{\inferrule*[Right={\scriptsize Def$\cap$ (11)}]{\inferrule*[Right={\scriptsize Def$\cap$
        (12)}]{\inferrule*[Right={\scriptsize distr (13)}]{\inferrule*[Right={\scriptsize Def$\cup$
            (14)}]{\inferrule*[Right={\scriptsize Def$\cap$ (15)}]{\inferrule*{}{y\!\in\!\textbf{T} \vdash y\!\in\!\textbf{T}}}{(y\!\in\!A \wedge y\!\in\!(B\!\cup\!C)) \vdash y\!\in\!\textbf{T}} }{(y\!\in\!A \wedge (y\!\in\!B \vee y\!\in\!C)) \vdash y\!\in\!\textbf{T}} }{(y\!\in\!A \wedge y\!\in\!B \vee y\!\in\!A \wedge y\!\in\!C) \vdash y\!\in\!\textbf{T}} }{(y\!\in\!A \wedge y\!\in\!B
         \vee y\!\in\!(A\!\cap\!C)) \vdash y\!\in\!\textbf{T}} }{(y\!\in\! (A\!\cap\!B) \vee y \in (A\!\cap\!C)) \vdash y\!\in\!\textbf{T}} }{(y\!\in\!\textbf{S}) \vdash y\!\in\!\textbf{T}}
 }{\vdash ((A \cap B) \cup (A \cap C)) \subseteq \textbf{T}}
}
{\vdash \underbrace{(A\!\cap\!(B\!\cup\!C))}_\textbf{T} = \underbrace{((A\!\cap\!B)\!\cup\!(A\!\cap\!C))}_\textbf{S}}
\end{mathpar}
\end{center}
\end{minipage}
} 
\caption{Assertion level proof for the statement $A \cap (B \cup  C)$ = $(A\cap B) \cup  (A\cap C)$
}
\label{alproof}
\vspace*{2em}
\end{figure*}}

\newcommand\figurebartlesherbert[1]{\begin{figure*}[#1]
\hspace*{-1.5em}\begin{tabular}{p{0.5\textwidth}p{0.45\textwidth}}
\subfigure[]{
\begin{minipage}{0.5\textwidth} 

\footnotesize

\noindent\begin{compactenum}
   \item In view of Definition 1.1.1, we [show] that the 
     sets $A \cap (B \cup C)$ and $(A \cap B) \cup (A \cap C)$ are equal. \steplabelA{13}
    $\left[ \mbox{First we show} \right]$  that $A \cap (B \cup C)$ 
     is a subset of $(A \cap B) \cup (A \cap C)$. \steplabelA{7} 
 $\left[ \mbox{Later we show} \right]$ $(A \cap B) \cup (A \cap C)$ is a subset of $A \cap (B \cup C)$.\steplabelA{12}
  \item Let $x$ be an element of $A\cap (B \cup C)$, \steplabelA{1}
  \item then $x \in A$ and $x \in B \cup C$. \steplabelA{2}
  \item This means that $x \in A$, and either $x \in B$ or $x \in C$.\steplabelA{3}
  \item  Hence we either have 
   (i) $x \in A$ and $x \in B$, or we have 
  (ii) $x \in A$ and $x \in C$.\steplabelA{4}
  \item Therefore, either $x \in A\cap B$ or $x \in A\cap C$,\steplabelA{5}
  \item so $x \in (A\cap B)\cup (A\cap C)$. \steplabelA{6}
  \item Conversely, let y be an element of $(A\cap B)\cup (A\cap C)$. \steplabelA{8}
  \item Then, either (iii) $y \in A \cap B$, or (iv) $y \in A \cap C$.\steplabelA{9}
  \item It follows that $y \in A$, and either $y \in B$ or $y \in C$. \steplabelA{10}
  \item Therefore, $y \in A$ and $y \in B \cup  C$, \steplabelA{11}
  \item so that $y \in A \cap (B \cup  C)$. \steplabelA{11} \\[-.5em]
\end{compactenum}
\end{minipage}
}
& 
\subfigure[]{
\begin{minipage}{0.45\textwidth}

\footnotesize

\noindent\begin{compactenum}
\item We show that   $((A\cap B)\cup (A\cap C) \subseteq A\cap B\cup C)$ and
  $(A\cap B\cup C \subseteq (A\cap B)\cup (A\cap C)) $
   ...because of definition of equality

\item We assume $x\in A\cap B\cup C$ and show $x\in (A\cap B)\cup (A\cap C)$

\item Therefore, $x\in A \wedge x\in B\cup C$

\item Therefore, $x\in A \wedge (x\in B\vee x\in C)$

\item Therefore, $(x$$\in$$A \wedge x$$\in$$B)\vee(x$$\in$$A\wedge x$$\in$$C)$

\item Therefore, $x\in A\cap B\vee x\in A\cap C$

\item We are done with the current part of the proof (i.e., to show that $x\in
  (A\cap B)\cup (A\cap C))$. [It remains to be shown that $(A\cap B)\cup
  (A\cap C)\subseteq A\cap B\cup C$] 

\item We assume $y\in (A\cap B)\cup (A\cap C)$ and show $y\in A\cap B\cup C$

\item Therefore, $y\in A\cap B\vee y\in A\cap C$

\item Therefore, $y\in A \wedge (y\in B\vee y\in C)$

\item Therefore, $y\in A \wedge y\in B\cup C$

\item This finishes the proof. Q.e.d. \\ 
\end{compactenum}
\end{minipage}
}
\end{tabular} \vspace*{-1em}

\caption{Comparison between (a) the (re-ordered) proof by Bartle and Sherbert
  \cite{BartleSherbert} and (b) the proof presentation generated with our rule set from the \OMEGA\ proof in Figure~\ref{alproof}}
\label{bartlesherbert}
\end{figure*}}

\newcommand\figurerules[1]{\begin{figure*}[#1] 
\hspace*{-1em}
\begin{tabular}{p{0.4\textwidth}p{0.7\textwidth}}
\subfigure[]{
\begin{minipage}{0.35\textwidth} 

\footnotesize

\begin{compactenum}[1)]
\item hypintro=1 $\wedge$ total$>1$ $\Rightarrow$ step-too-big
\item $\cup$-Defn$\in${}$\{1,2\}${}$\wedge${}$\cap$-Defn$\in${}$\{1,2\}$ $\Rightarrow$ step-too-big
\item $\cap$-Defn$<\!3$ $\wedge$ $\cup$-Defn=0 $\wedge$ masteredconceptsunique=1
$\wedge$ unmasteredconceptsunique=0 $\Rightarrow$ step-too-small 

\item total$<$2 $\wedge$ verb=true $\Rightarrow$ step-too-small
\item masteredconceptsunique$<$3 $\wedge$ unmasteredconceptsunique=0 $\wedge$ verb=true $\Rightarrow$ step-too-small
\item equalitydefn$>$0 $\wedge$ verb=false $\Rightarrow$ step-too-big

\item $\underbar{\ }$ $\Rightarrow$ step-appropriate \\[-.5em]
\end{compactenum}
\end{minipage}
}
&
\subfigure[]{
\begin{minipage}{0.55\textwidth}
\footnotesize
\begin{compactenum}[1)]
\item conceptsunique\,$\in${}$\{0,1\}$ $\wedge$ equalitydefn=0 $\wedge$ verb=true $\Rightarrow$ step-too-small
\item hypintro=0 $\wedge$ equalitydefn=0 $\wedge$ $\cup$-Defn=0 $\wedge$ verb=true $\Rightarrow$ step-too-small
\item conceptsunique $\in${}$\{2,3,4\}$ $\wedge$ $\cup$-Defn $\in${}$\{1,2,3\}$  $\Rightarrow$  step-too-big
\item hypintro $\in${}$\{1,2,3,4\}$ $\wedge$ conceptsunique $\in${}$\{2,3,4\}$ $\Rightarrow$  step-too-big
\item unmasteredconceptsunique=0 $\wedge$ total $\in${}$\{0,1,2\}$ $\cap$-Defn $\in${}$\{1,2\}$ $\wedge$ close=false $\Rightarrow$ step-too-small
\item equalitydefn $\in${}$\{1,2\}$ $\wedge$verb=false $\Rightarrow$  step-too-big
\item equalitydefn$\in${}$\{1,2\}$ $\wedge$ verb=true $\Rightarrow$ step-appropriate
\item equalitydefn=0 $\wedge$ verb=false $\Rightarrow$ step-appropriate

\item $\underbar{\ }$ $\Rightarrow$ step-appropriate \\
\end{compactenum}
\end{minipage}
}
\end{tabular} \vspace*{-1.5em}

  \caption{\ Rule sets employed in the running example: \
(a)\,~rule set generated by hand, \ 
(b)\,~rule set generated using C5.0 
    (ordered by the rules' confidence values)}
  \label{rules}
\vspace*{2em}
\end{figure*}}

\newcommand\figurerunningtwo[1]{\begin{figure}[#1] 
\footnotesize
\begin{compactenum}
\item We show that   $((A\cap B)\vee (A\cap C)\subseteq A\cap B\vee C)$ and $(A\cap B\vee C\subseteq (A\cap B)\vee (A\cap C))$ 
...because of definition of equality
\item We assume $x\in A\cap B\vee C$ and show $x\in (A\cap B)\vee (A\cap C)$
...because of definition of subset
\item Therefore, $x\in A \wedge x\in B\vee C$
...because of definition of intersection
\item Therefore, $x\in A \wedge (x\in B\vee x\in C)$
...because of definition of union
\item Therefore, $x\in A \wedge x\in B\vee x\in A \wedge x\in C$
...because of logics
\item Therefore, $x\in A\cap B\vee x\in A\cap C$
...because of definition of intersection
... similarly to step nr. 3 
\item We are done with the current part of the proof (i.e., to show that $x\in (A\cap B)\vee (A\cap C))$. It remains to be shown that $(A\cap B)\vee (A\cap C)\subseteq A\cap B\vee C$. ... because of definition of union.
\item We assume $y\in (A\cap B)\vee (A\cap C)$ and show $y\in A\cap B\vee C$
...because of definition of subset
\item Therefore, $y\in A\cap B\vee y\in A\cap C$
...because of definition of union
\item Therefore, $y\in A \wedge y\in B \vee y\in A \wedge y\in C$
...because of definition of intersection
... similarly to step nr. 3 
\item Therefore, $y\in A \wedge (y\in B\vee y\in C)$
...because of logics
\item Therefore, $y\in A \wedge y\in B\vee C$
...because of definition of union
\item This finishes the proof. Q.e.d. ...because of definition of intersection
\end{compactenum}

\caption{The assertion level proof in Figure~\ref{alproof}
presented according to the rule set from Figure~\ref{rules2}}
\label{running2}
\vspace{2ex}
\end{figure}}

\newcommand\maslong{mathematics assistance system}
\newcommand\putinquotes[1]{``#1''}
\newcommand\noitem{\vspace{-1.0\itemsep}}
\newcommand\OMEGA{{$\Omega${\sc mega}}}
\def\dialog{{\sc Dialog}\xspace}

\newcommand{\CORE}{\textsc{CoRe}\xspace}
\newcommand\steplabelA[1]{\color{black}{\setlength{\fboxsep}{0\baselineskip}\colorbox{lightgray}{\setlength{\fboxsep}{0.1\baselineskip}\framebox{#1}}}\color{black}{}}

\begin{document}
\makecover
\maketitle
\begin{abstract}%
When mathematicians present proofs they usually adapt their
explanations to their didactic goals and to the (assumed) knowledge
of their addressees.  Modern automated theorem provers, in contrast,
present proofs usually at a fixed level of detail (also called
granularity).  Often these presentations are neither intended nor
suitable for human 
use. 
A challenge therefore is to develop
user- and goal-adaptive proof presentation techniques that obey
common mathematical practice. We present a flexible and adaptive
approach to proof presentation that exploits machine learning
techniques to extract a  model of the specific granularity of 
proof examples and employs this model for the automated generation 
of further proofs at an adapted level of granularity.%
\end{abstract}

\vfill\pagebreak

\section{Introduction}
\figurebert{t}%
A key capability trained by students in mathematics and the formal
sciences is the ability to conduct rigorous arguments and proofs and
to present them.  Thereby, proof presentation is usually highly
adaptive as didactic goals and (assumed) knowledge of the addressee
are taken into consideration. Modern automated theorem proving
systems, however, do often not sufficiently address this common
mathematical practice.  They typically generate and present proofs
using very fine-grained and machine-oriented calculi. While some
theorem proving systems exist 
---~amongst them prominent interactive theorem provers~--- 
that provide means for human-oriented proof
presentations (e.g. proof presentation modules in   
Coq~\cite{Thery95extractingtext}, Isabelle~\cite{simons-isabelle97}
 and Theorema~\cite{RISC2674}),
the challenge of supporting user- and goal-adapted
proof presentations has been widely neglected in the past. This
constitutes an unfortunate gap, in particular since mathematics and
the formal sciences are increasingly targeted as promising application
areas for intelligent tutoring \nolinebreak systems. 
In this paper we present a flexible and adaptive approach to proof presentation
that exploits machine learning techniques to extract a model of the specific 
granularity of given proof examples, 
and that subsequently employs this model for the automated
generation of further proofs at an adapted level of granularity.
Our research has its roots in the collaborative \dialog project
\cite{DBLP:conf/cogsys/BenzmullerHKPSW05} in which we developed means
to employ the proof assistant \OMEGA\
\cite{DBLP:journals/japll/SiekmannBA06} for the dialog-based teaching
of mathematical proofs.  In \dialog we have considered a dynamic
approach: Instead of guiding the student along a pre-defined path
towards a solution, we support the dynamic exploration of proofs,
using automated proof search.  This presupposes the development of
techniques to adequately model the proofs a student is supposed to
learn.
Inference steps in \OMEGA\ are implemented via an \emph{assertion
  application} mechanism~\cite{Dietrich-06-a}, which is based upon
Serge Autexier's \CORE calculus~\cite{DBLP:conf/cade/Autexier05} as
its logical kernel. \ 
In assertion level proofs, all inference steps are justified by a
mathematical fact, such as definitions, theorems and lemmas, but not
by steps of a purely technical nature such as structural
decompositions, as required, for example, in natural deduction or
sequent calculi. 

The development of the dialog system prototype was guided by empirical
studies using a mock-up of the \dialog system \cite{lrec06}. 
One research challenge that educed out of the
experiments is the question of judging the appropriate step size of
proof steps (in the context of tutoring), also referred to as the
\emph{granularity} of mathematical proofs.  Even in introductory textbooks
in mathematics, intermediate proof steps are skipped, when this seems
appropriate.  An example is the elementary proof in basic set theory
reproduced in Figure~\ref{bartlesherbertproof}\@. 
\pagebreak
Whereas most of the
proof steps consist of the application of exactly one mathematical
fact (in this case, a definition or a lemma, such as the
distributivity of \emph{and} over \emph{or}), the step from assertion \steplabelA{9}
to assertion \steplabelA{10} suggests the application of several inference steps
at once, namely the application of the definition of $\cap$ twice, and
then using the distributivity of \emph{and} 
over  \emph{or}.  
\par
\begin{wrapfigure}[9]{r}{8.8cm}
\small
\textbf{student:}  $ (x,y) \in (R \circ S)^{-1} $ \\
\textbf{tutor:} \texttt{Now~try~to~draw~inferences~from~that!}\\
\begin{tabular}{|l|l|l|} \hline correct & appropriate & relevant \\  \hline \end{tabular} \vspace*{4ex} \\
\textbf{student:}  $(x,y) \in S^{-1} \circ R^{-1} $ \\
\textbf{tutor:} \texttt{One cannot directly deduce that.}\\
\begin{tabular}{|l|l|l|} \hline correct & too coarse-grained & relevant\\ \hline \end{tabular}
\label{granularityproblem}
\end{wrapfigure}
Similar observations were made in the empirical studies within the
\dialog project.  In these studies the tutors who helped to simulate
the dialog system identified limits for how many inference steps are
to be allowed at once.
An example for an inacceptably large student step that was rejected by the 
tutor is presented to the right.

The idea to represent proofs at different levels 
of detail was incorporated 
into \OMEGA\ as a hierarchically organized proof data structure \cite{C19}. \
The proof explanation system P.rex \cite{Fiedler:ijcar01} implemented
the idea to generate adapted proof presentations by moving up or down
these layers on request. \ 
Alas, though the proofs at different levels
of detail can be handled by the \OMEGA\ system, the problem remains of
how to identify a particular level of granularity and how to ensure
that this level of granularity is appropriate. \
This observation also applies to the Edinburgh HiProofs system~
\cite{hiproofs2006}.

Autexier and Fiedler have proposed one particular level of granularity
\cite{DBLP:conf/mkm/AutexierF05}, which they call
\emph{what-you-need-is-what-you-stated granularity}. \
Based on the
assertion level inference mechanism in \OMEGA, 
they also developed a proof checking mechanism for this level. \
In \nolinebreak brief, their notion of
granularity refers to such assertion level proofs, where all assertion
level inference steps are spelled out explicitly and refer only to
facts readily available from the assertions or the previous inference
steps. \
However, they conclude that even the simple proof in
Figure~\ref{bartlesherbertproof} 
does not comply with their level of granularity,
since the proof is missing some details.
 
This paper presents in Section~\ref{model} an adaptive framework to
model proof granularity. \
This framework has been implemented as an
extension of the \OMEGA\ proof assistant and 
it \nolinebreak is used to generate
proof presentations at specific granularity levels of interest. \
In Section~\ref{case} we illustrate how our framework captures the
granularity of our running example proof in
Figure~\ref{bartlesherbertproof}\@. \  
Models for granularity can be learned
in our framework from samples, for which we employ standard machine
learning techniques, as demonstrated in Section~\ref{learning}\@. 

\vfill\pagebreak
\section{An Adaptive Model for Granularity}
\label{model}

We treat the granularity problem as a classification task: given a
proof step, representing one or several assertion applications, we
judge it as either \emph{appropriate}, \emph{too big} or \emph{too
  small}. As our feature space we employ several mathematical and
logical aspects of proof steps, but also aspects of cognitive
nature. For example, we keep track of the background knowledge of the
user in a student model.

We illustrate our approach with an example proof step in
Figure~\ref{bartlesherbertproof}: \
\steplabelA{10} is derived from \nolinebreak \steplabelA{9} 
by applying the definition of $\cap$ twice, and then
using the distributivity of \emph{and} over \emph{or}.  In this step
(which corresponds to multiple assertion level inference steps) we
make the following observations:\begin{enumerate}\noitem\item[(i)] 
involved are two concepts: def. of $\cap$ and
  distributivity of \emph{and} over \emph{or}, 
\noitem\item[(ii)] 
the total number of assertion applications is three, 
\noitem\item[(iii)] 
all involved concepts have been previously applied
  in the proof,
\noitem\item[(iv)]
all manipulations apply to a common part in \steplabelA{9},
\noitem\item[(v)] 
the names of the applied concepts are not explicitly mentioned, and 
\noitem\item[(vi)]
two of the assertion applications belong to \emph{naive set theory}
    (def. of $\cap$) and one of them relates
    to the domain of propositional logic (distributivity).
\noitem\end{enumerate}
%
\figurealproof{t}%
These observations can be represented as a feature vector,\footnote
{Currently, we use around twenty features  
 which are domain-independent, 
 plus an indicator feature for each definition or lemma, 
 and one indicator feature for each theory.} 
where, in our
example, the feature ``distinct concepts'' receives a value of 
``\hskip.009em2\hskip.009em'', and so forth. \ 
We express our models for classifying granularity as rule sets, which
associate specific combinations of feature values to a corresponding granularity
verdict (``appropriate'', ``too big'' or ``too small''). \
These rule sets may be
hand-authored by an expert or they may be learned 
from empirical data as we show in Section~\ref{learning}\@. \ 
Our algorithm for granularity-adapted proof presentation takes two arguments, a
granularity rule set and an assertion level proof\,\footnote
{Our approach is not restricted to assertion level proofs 
 and is also applicable to other calculi. \
 However, in mathematics education we consider single assertion level proof
  steps as the finest granularity level of interest. \
 We gained evidence for this
 choice from the empirical investigations in the \dialog project 
 \mbox{(cf.\ \cite{DBLP:conf/cogsys/BenzmullerHKPSW05} and \cite{lrec06})}.
} 
as generated by \OMEGA\@. \ 
Figure~\ref{alproof}\, 
shows the assertion level proof generated by \OMEGA\ for our
running example; this proof is represented as a tree (or acyclic graph) in
sequent-style notation and the proof steps are ordered. \
Currently we 
only consider plain assertion level proofs, and do not assume 
any prior hierarchical structure or choices between proof alternatives 
(as possible in \OMEGA). \
Our
algorithm performs an incremental categorization of steps in the proof 
tree 
 (where 
$n=0,\ldots,k$ denotes the ordered proof steps in the tree; initially $n$
is $1$):

\begin{quote}
\noindent\texttt{\bf while} \ there exists a proof step $n$ \ \texttt{\bf do}\\ 
evaluate the granularity of the compound proof step $n$ (i.e., the proof step 
consisting of all assertion level inferences performed after the last step 
labeled 
\putinquotes{appropriate with explanation} 
or
\putinquotes{appropriate without explanation} 
--- or the beginning
of the proof, if none exists yet) 
with the given rule set under each of the following two assumptions: 
(i) assuming that the involved concepts are mentioned in the 
presentation of the step (an \emph{explanation}), 
and (ii) assuming that only the resulting formula is displayed.
\begin{compactenum}
\item \sloppy\texttt{\bf if} \ $n$ is 
{appropriate} with explanation\\
\texttt{\bf then} \ label $n$ as 
\putinquotes{appropriate with explanation};\, 
\mbox{set $n:=n$$+$$1$}; 
\item \texttt{\bf if} \ 
$n$ is 
{too small} with explanation, 
but 
{appropriate} without explanation
\\\texttt{\bf then} \
label $n$ as 
\putinquotes{appropriate without explanation};\, 
set $n:=n$$+$$1$; 
\item \texttt{\bf if} \ 
$n$ is 
{too small} both with and without explanation
\\\texttt{\bf then} \
label $n$ as 
\putinquotes{too small};\, 
set $n:=n$$+$$1$; 
\item \texttt{\bf if} \ $n$ is 
{too big} 
\\\texttt{\bf  then} \ 
label $n$$-$$1$ as  
\putinquotes{appropriate without explanation} 
(i.e.\ \nolinebreak consider the previous step as appropriate), \
unless $n$$-$$1$ is labeled 
\putinquotes{appropriate with explanation} 
or
\putinquotes{appropriate without explanation} 
already or $n$ is the first step in the proof 
(in this special case label $n$ as 
\putinquotes{appropriate with explanation} and set $n:=n$$+$$1$). 
\end{compactenum}
\texttt{\bf od}
\end{quote}
We thereby obtain a proof tree with labeled steps (or labeled nodes) which
differentiates between those nodes that are categorized as appropriate for
presentation and those which are considered too
fine-grained. \
Proof presentations are generated 
by walking through the tree,\footnote
{In case of several 
 branches, a choice is possible which subtree to present first, a question 
 which we do not address in this paper.%
}
skipping the steps labeled \emph{too small}.\footnote
{Even though the intermediate steps which are \emph{too small} are 
 withheld, the presentation of the 
 output step reflects the results of all intermittent assertion applications, 
 since we include the names of all involved 
 concepts whenever a (compound) step is appropriate with explanation.}



When modeling granularity as a categorization problem, 
we have to test the hypothesis that the combination of features 
we devise is useful for the classification task. \ 
I.e., we have to determine whether 
steps within a class (i.e. ``appropriate'', ``too big'' and ``too small'') can
indeed be fruitfully characterized by specific combinations of feature values,
and distinguished from the feature values that characterize the two other
classes.  \
Our methodology for evaluation of this hypothesis 
consists in case studies and in
empirical evaluations with mathematics tutors. \
This is exemplified in the
following two sections.\vfill\pagebreak

\figurebartlesherbert{!h}%

\vfill

\figurerules{!h}%

\vfill\clearpage

\section{Case Study}\label{case}
In this section, we exemplarily model the step size of the textbook proof
 in Figure~\ref{bartlesherbertproof}\@. \
Starting point for the automated generation of our proof presentations 
are assertion level proofs in the \maslong\ \OMEGA\@.
The basic assertion level proof, assuming the basic definitions in
naive set theory, is presented in Figure~\ref{alproof} as a sequent style proof tree. 

This proof consists of fifteen assertion level inference applications,
which refer to the definitions of equality, subset, union and
intersection as well as the concept of distributivity. \
Notice that the
 proof in Figure~\ref{bartlesherbertproof} (taken from the textbook 
Bartle \& Sherbert~\cite{BartleSherbert}) 
starts (in statement~\steplabelA{1}) with the assumption 
that an element $x$ is in the set $A\cap{}(B\cup{}C)$. \
The intention is to show the subset relation
$A\cap(B\cup{}C)\subseteq{}(A\cap{}B)\cup{}(A\cap{}C)$. \ 
However, this
is not explicitly revealed until step \steplabelA{6}, when this part
of the proof is already finished. \
The same style of 
\emph{delayed} justification for prior steps is employed towards the
end of the proof, where statements \steplabelA{12} and \steplabelA{13}
justify (or recapitulate) the preceding proof. \
It must be questioned whether this style of presentation,
where the motivation
for some of the steps (such as the above assumption)
is only presented in retrospective (when the assumption is discharged), 
is still the most effective one for instructing students in our times. \
This style originated in former centuries, when the general task of the
apprentice was to figure out the reason behind the procedures of
his technically highly competent master with often poor teaching skills. \

Thus, for the modeling of step size, we consider
a re-ordered variant of the steps in Figure~\ref{bartlesherbertproof},
which is displayed in Figure~\ref{bartlesherbert}~(a).\footnote
{Note that
  step (1) in the re-ordered proof corresponds to the statements
  \steplabelA{7}, \steplabelA{12} and \steplabelA{13} in the original
  proof which jointly apply the concept of set equality.} \
We now generate a proof presentation which matches the step size of
the twelve steps in the original proof, skipping intermediate proof
steps according to our feature-based granularity model. \
Figure~\ref{rules} shows two sample rule sets 
which both lead to the proof presentation  
in Figure~\ref{bartlesherbert}~(b)\@. \
The rule set in Figure~\ref{rules}~(a) was generated by hand, 
whereas the rule set in Figure~\ref{rules}~(b) was generated with the 
help of the C5.0 data mining tool \cite{rulequest2007}.\footnote
{The sample proof was used to fit the rule set to it. \
 All steps in the 
 sample proof were provided as \emph{appropriate}, all 
 intermediate assertion level steps were labeled as \emph{too-small}, 
 and always the next bigger step to each step in the original proof was 
 provided as an example for a \emph{too big} step. \
 Care was taken 
 that the default rule of the generated 
 rule set is of class \emph{appropriate} (which was 
 achieved via the cost function), so that the rule set better transfers to 
 other domains. \
 Otherwise, in case the default class was \emph{too small}, 
 and the examined proof steps were sufficiently different from 
 the generating sample (and thus failed to match the non-default rules), 
 the resulting proof presentation would be excessively short.}   

The feature \emph{hypintro} indicates whether a (multi-inference) proof step
introduces a new hypothesis, and \emph{close} indicates whether 
a branch of the proof has been finished. \
The feature \emph{total} counts the
number of assertion level inferences within one (multi-inference) step. \
Furthermore, the features \emph{masteredconceptsunique} and
\emph{unmasteredconceptsunique} indicate how many of the employed
concepts (if any) are supposed to be mastered or unmastered by the
user according to a very basic student model (which is updated in the course of
the proof). \
Furthermore, the occurrences of particular defined notions
are counted (via the features $\cap$-Defn, $\cup$-Defn, equalitydefn). \
For example, 
the  first rule in Figure~\ref{rules}~(a) can be interpreted as 
``If a step introduces a new hypothesis into the proof, 
  and consists of more than one assertion level inference rule, 
  it is considered too big.'' \
Note that rules~4--6 
in Figure~\ref{rules}~(a) express the relation between the appropriateness of 
steps and whether the employed concepts are mentioned verbally 
(feature \emph{verb}). \
Rule 6 has the effect of enforcing that the use of the 
definition of equality is always explicitly 
mentioned (as in step 1.\ in Fig~\ref{bartlesherbert}~(b))\@. \ 
All other cases, which are not covered by the previous rules, are
subject to a default rule. \
Rules are ordered by utility for conflict resolution.

The generated proof presentation in Figure~\ref{bartlesherbert} (b) consists, 
similarly to the proof in Figure~\ref{bartlesherbert} (a), of 
twelve steps. \
The three assertion 
level steps (11), (12) and (13) are combined into one single step 
from (9) to (10) in 
Figure~\ref{bartlesherbert} (b)\@. \
Natural language is produced via simple patterns. \
(A more exciting natural language generation 
is possible with Fiedler's mechanisms \nolinebreak\cite{Fiedler:ijcar01}, \
but this is not the subject of this paper.)






The rule sets in Figure~\ref{rules} can be successfully reused for other examples in the domains as well. \ 
In  Figure~\ref{newproof}, we present the resulting proof presentation when applying 
the rule set in Figure~\ref{rules} (a) to a different proof exercise,
namely a proof of the theorem 
$$(A \cap B) \backslash C = A \cap (B \backslash C).$$ 

\vfill\vfill\vfill

\begin{figure*}[!h] \footnotesize
\begin{compactenum}
\item  We show that   $((A \cap B)\backslash C \subseteq A\cap B\backslash C$) and $(A \cap
  B\backslash C \subseteq (A\cap B)\backslash C)$  ...because of definition of equality
\item We assume $x\in A\cap B\backslash C$ and show $x\in (A\cap B)\backslash C$ 
\item Therefore, $x\in A \wedge  x\in B\backslash C$ 
\item Therefore, $x\in A \wedge  x\in B \wedge  \neg (x\in C)$  
\item  We are done with the current part of the proof (i.e., to show that
  $x\in (A\cap B)\backslash C)$. It remains to be shown that $(A\cap  B \backslash C \subseteq A \cap B \backslash C$.

\item  We assume $y\in (A\cap B)\backslash C$ and show $y\in A\cap B\backslash
  C$ 
\item Therefore, $y\in A \wedge  y\in B \wedge  \neg(y\in C)$ 
similarly to steps nr. (3 4) 
\item This finishes the proof. Q.E.D. 
... similarly to step nr. 7 
\end{compactenum}
\caption{Sample proof presentation generated via the rule set in 
Figure~\ref{rules} (a) for the theorem $(A \cap B) \backslash C = A \cap (B \backslash C)$}
\label{newproof}
\vspace*{2em}
\end{figure*}

\vfill

\begin{figure}[!h]
\footnotesize
\begin{verbatim}
PART decision list
------------------
total <= 2 AND total > 0 AND parapos <= 0: appropriate (85.0/4.0)

total <= 2 AND unmasteredconceptsunique <= 0: step-too-small (11.0/2.0)

parapos <= 0 AND samesub <= 0: step-too-big (22.0/5.0)

unmasteredconceptsunique <= 1 AND hypintro <= 0: appropriate (9.0)
: step-too-big (8.0/2.0)
\end{verbatim}\vspace*{-3ex}
\caption{Empirically learned rule set. The feature \emph{parapos} indicates whether an inference has been applied only once in a proof situation where it could have been applied twice, in the same direction. The feature \emph{samesub} indicates whether all inference applications within a (multi-inference) step apply to the same formula (and the same subparts thereof).}
\label{rules2}
\end{figure}

\cleardoublepage

\section{Learning from Empirical Data}
\label{learning}

Classification problems are a well-investigated topic in the 
machine learning community. There exist off-the-shelf 
tools that allow to learn classifiers  
(like our rule sets) 
from annotated examples (\emph{supervised} learning). 
In our case, an expert annotates proof steps with the labels 
\emph{appropriate}, \emph{too small} or \emph{too big}. 
Representing the proof steps in \OMEGA\ has the advantage that 
all the features of a particular proof step are computed in 
the background, and combined automatically with the expert's judgments 
as training instances for the learning algorithm. 
Currently, our algorithm calls the C5.0 data mining tools \cite{rulequest2007,DBLP:books/mk/Quinlan93} 
---~which support the learning of decision trees and of rule sets~--- 
to obtain classifiers for granularity. 

As part of an ongoing evaluation, we have conducted a 
study where a mathematician (with tutoring experience) judged the granularity 
of 135~proof steps. \
These steps were presented to him via an \OMEGA-assisted 
environment which computed the feature values for granularity classification 
in the background. \
The step size of proof steps presented to the expert 
was randomized, such that each presented step corresponded to 
one, two, or three assertion level inference steps. \
The presented proofs belonged to one exercise in naive set theory 
and three different exercises about relations. \ 
We evaluated rule learning using C5.0 on our sample 
using 10 fold cross validation, 
which resulted in a mean percentage of correct classification of 
84.6\%, and $\kappa=0.62$. \
We also used the PART classifier~\cite{Frank98generatingaccurate} included 
in the Weka suite\footnote{\url{http://www.cs.waikato.ac.nz/~ml/weka/}}, 
which is inspired by Quinlan's C4.5. \ 
After we excluded some of the attributes (in particular 
those that refer to the use of specific concepts, i.e., 
Def.\ \nolinebreak of \nolinebreak$\cap$, Def.\ of $\circ$, etc.), 
PART  achieved 86.7\% of correctly classified instances 
in stratified cross validation ($\kappa$=0.68). \
Apparently, removal of the most domain-specific attributes 
prevented the algorithm from overfitting. \
The resulting rule set is presented 
in Figure~\ref{rules2}\@. 

The feature \emph{parapos} indicates whether an inference has been applied only once in a proof situation where it could have been applied twice, in the same direction. \
The feature \emph{samesub} indicates whether all inference applications within a (multi-inference) step apply to the same formula 
(and the same subparts thereof). \
When applied to our running example, 
we obtain the proof presentation as shown in Figure~\ref{running2}\@.%
\figurerunningtwo{ht}

To compare the rule-based 
classifiers with support vector machines, we applied SMO~\cite{platt98} on our data, 
resulting in 83.0\% correctness and $\kappa$=0.57 in stratified 
cross validation, which is a similar performance to C5.0.

\vfill\cleardoublepage
\section{Conclusion}
\label{conclusion}
Granularity has been a challenge in AI for decades
\cite{DBLP:conf/ijcai/Hobbs85,Mccalla92granularityhierarchies}. Here
we have focused on adaptive proof granularity, which we treat as a
classification problem.  We model different levels of granularity
using rule sets, which can be hand coded or learned from sample
proofs. 

As a case study, we have 
formulated the granularity level of the 
proof in Figure~\ref{bartlesherbertproof} 
from the textbook \cite{BartleSherbert} 
as a rule set in our classification-based approach. 
Classifiers 
are applied dynamically to each proof step, thus taking into 
account changeable information such as the user's familiarity with 
the involved concepts. Using assertion level proofs as the basis 
for our approach has the additional advantage that the relevant 
information for the classification task (e.g., the concept names) 
is easily read off the proofs. This also eases the generation 
of natural language proof output in general.




Future work consists in empirical evaluations of the learning approach ---
to address the following questions: 
\begin{enumerate}
\item[(i)] what are the most useful
features for judging granularity, and are they different among distinct experts and domains, 
\item[(ii)] what is the interrater reliability among different experts and the corresponding 
classifiers generated by learning in our framework?
\end{enumerate}
The resulting corpora of 
annotated proof steps and generated classifiers can then be used to 
evaluate the appropriateness of the proof presentations generated by our system.

\section*{Acknowledgements}

We thank Claus-Peter Wirth for his
thorough and helpful review of the paper. 
Furthermore, we thank Erica Melis 
and her ActiveMath group for valuable institutional 
and intellectual support of this work.

\bibliographystyle{abbrv}

\end{document}